\documentclass{article}

\usepackage{arxiv}

\usepackage[utf8]{inputenc} % allow utf-8 input
\usepackage[T1]{fontenc}    % use 8-bit T1 fonts
\usepackage{url}            % simple URL typesetting
\usepackage{graphicx}
\usepackage{booktabs}       % professional-quality tables
\usepackage{amsfonts}       % blackboard math symbols
\usepackage{microtype}      % microtypography

\usepackage{amsmath,amsfonts,bm}

\def\eqref#1{equation~\ref{#1}}
\def\1{\bm{1}}

\DeclareMathAlphabet{\mathsfit}{\encodingdefault}{\sfdefault}{m}{sl}
\SetMathAlphabet{\mathsfit}{bold}{\encodingdefault}{\sfdefault}{bx}{n}

\usepackage{algorithm}
\usepackage{algorithmic}
\usepackage[table]{xcolor}
\usepackage{caption}
\usepackage{multirow}
\usepackage{amsmath}

\usepackage{natbib}
\setcitestyle{aysep={}}
\let\cite\citep

\usepackage{caption}
\usepackage[colorlinks=true, allcolors=black]{hyperref}

\title{Can Vision-Language Models Synthesize Executable Game-Playing Programs?}

\author{
    \textbf{Yao Tong}\thanks{Equal contribution.}\\
    Hangzhou Institute for Advanced Study\\
    University of Chinese Academy of Sciences\\
    Hangzhou, China\\
    \texttt{tongyao261@mails.ucas.ac.cn} \\
    \And
    \textbf{Jiaxin Qi}\footnotemark[1]\\
    Computer Network Information Center\\
    Chinese Academy of Sciences\\
    Beijing, China\\
    \texttt{jxqi@cnic.cn} \\
    \And
    \textbf{Dingyu Zhang}\\
    Hangzhou Institute for Advanced Study\\
    University of Chinese Academy of Sciences\\
    Hangzhou, China\\
    \texttt{zhangdingyu26@mails.ucas.ac.cn} \\
    \And
    \textbf{Jianqiang Huang}\thanks{Corresponding author.}\\
    Computer Network Information Center\\
    Chinese Academy of Sciences\\
    Beijing, China\\
    \texttt{jqhuang@cnic.cn} \\
}

\renewcommand{\headeright}{Preprint}
\renewcommand{\undertitle}{Preprint}
\renewcommand{\shorttitle}{Can VLMs Synthesize Game-Playing Programs?}

\hypersetup{
pdftitle={Can Vision-Language Models Synthesize Executable Game-Playing Programs?},
pdfsubject={cs.AI, cs.CV, cs.CL},
pdfauthor={Yao Tong, Jiaxin Qi, Dingyu Zhang, Jianqiang Huang},
pdfkeywords={Vision-Language Models, Benchmark, Program Synthesis, Game-Playing Agents, Code Generation},
}

\begin{document}

\maketitle

% =====================================================================
% Abstract
% =====================================================================
\begin{abstract}
Recent advances in vision-language models (VLMs) have endowed multimodal agents with increasingly strong visual perception and long-horizon reasoning capabilities, making VLM-based autonomous game playing feasible. Existing game benchmarks primarily evaluate whether an agent can select appropriate actions from the current visual state across different scenarios and games. However, most of them adopt a step-by-step interaction paradigm where the current visual observation is sent to the agent before each action. In practical deployment, this paradigm introduces two fundamental limitations: the game loop must wait for model inference, making real-time interaction impractical, and repeated VLM calls incur substantial cost. Therefore, we propose PlayScriptBench, a script-based game benchmark that evaluates whether VLMs can synthesize an executable game-playing program from game states. 
This setting jointly evaluates visual grounding, understanding of game dynamics, long-horizon strategic reasoning, and executable program synthesis, providing a more comprehensive assessment of game-playing agents. To support more tractable evaluation for current general-purpose VLMs, we further introduce a four-stage pipeline --- Explore, Perception, API Construction, and Strategy Composition --- that decomposes controller synthesis into game exploration, state perception, interface construction, and strategic control, enabling agents to progressively construct an executable program. Experiments on 16 games spanning 4 genres show that even strong VLMs struggle on our benchmark, revealing a substantial gap between passive visual game understanding and executable closed-loop game control. These results establish script-based controller synthesis as a complementary direction for benchmarking VLM-based game-playing agents.
\end{abstract}

% =====================================================================
% Introduction
% =====================================================================
\section{Introduction}

\begin{figure}[t]
\centering
\includegraphics[width=0.7\columnwidth]{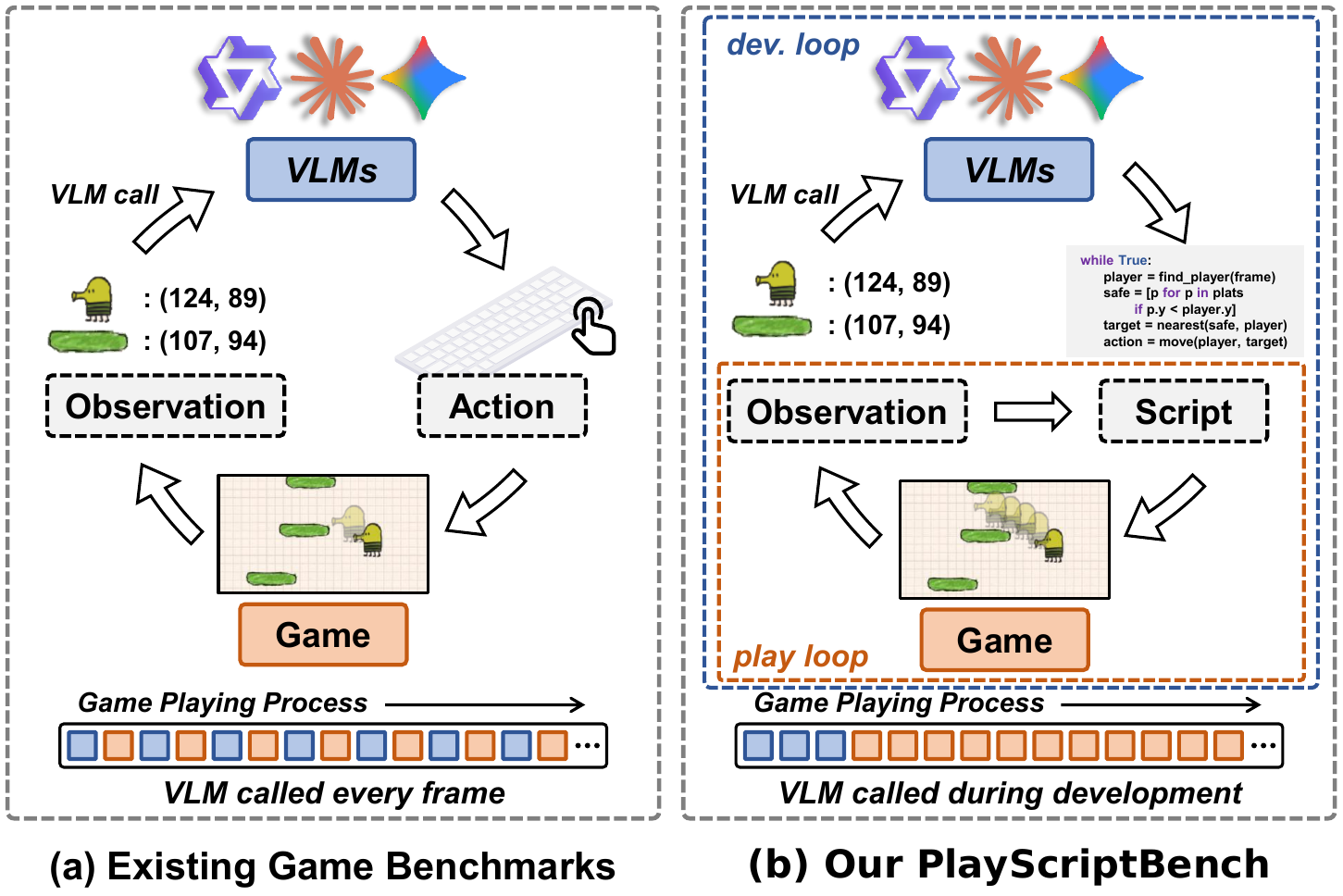}
\caption{Comparison of evaluation protocols. \textbf{Left:} Existing game-agent benchmarks invoke the VLM inside the observation--action loop: every frame, the model receives a screenshot and predicts an action. \textbf{Right:} PlayScriptBench inverts the protocol---the VLM is queried only once during a development phase to synthesize an executable controller, which then runs autonomously.}
\label{fig:protocol_comparison}
\end{figure}

Recent advances in vision-language models (VLMs) have endowed multimodal agents with increasingly strong visual perception and long-horizon reasoning capabilities, making VLM-based autonomous game playing feasible~\cite{ouyangGameWorldStandardizedVerifiable2026}. A parallel thread---visual program synthesis (VPS)~\cite{surisViperGPTVisualInference,guptaVisualProgrammingCompositional2022,luChameleonPlugandPlayCompositional2023}---demonstrates that LLMs can generate executable Python programs from visual inputs, grounding language in composable code modules for visual reasoning. Video games provide a natural testbed for both capabilities: they demand fine-grained visual understanding of dynamic scenes and sustained planning to achieve goals, making them an ideal benchmark for evaluating whether VLMs can act autonomously in visual environments.

Yet the way current game benchmarks are designed constrains how VLMs can participate in games (Figure~\ref{fig:protocol_comparison}, left). Whether the target is arcade games~\cite{zhangVideoGameBenchCanVisionLanguage2026}, open-world sandboxes~\cite{ouyangGameWorldStandardizedVerifiable2026}, or puzzle environments~\cite{ouyangGameWorldStandardizedVerifiable2026}, the per-frame VLM-in-the-loop design is pervasive: at every step, the model receives the current observation, predicts the next action, and executes it, before repeating the cycle from scratch. While practical mitigations---frame skip, observation caching, or asynchronous inference---can reduce per-step cost and latency, the paradigm still ties inference budget to game duration and limits the controller to the model's instantaneous reasoning horizon. PlayScriptBench inverts this design: instead of querying the VLM at every frame, we ask it to synthesize a standalone executable controller that runs autonomously at native frame rate. This decouples inference cost from play time. We note that this paradigm is complementary to, not a replacement for, per-frame inference: script synthesis compresses visual understanding into persistent code and is most informative for deterministic, bounded-state games, whereas per-frame inference excels at online reactivity in open-world or stochastic environments.

\begin{figure*}[t]
\centering
\includegraphics[width=0.95\textwidth]{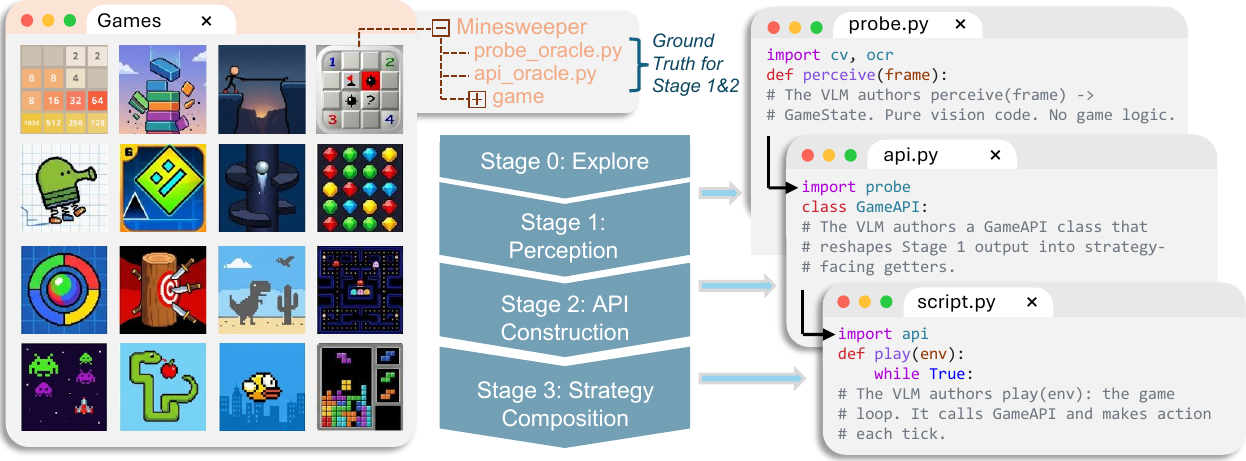}
\caption{Overview of the PlayScriptBench: the four-stage pipeline and the 16 HTML5 games.}
\label{fig:pipeline}
\end{figure*}

We introduce PlayScriptBench as a \emph{complementary} benchmark that targets this script-synthesis regime. PlayScriptBench is a script-based game benchmark that evaluates whether VLMs can synthesize an executable game-playing program from game states, focusing on the class of deterministic, bounded-state games where a static controller can in principle play reliably. Visual program synthesis serves as the evaluation \emph{means}; game behavioral performance is the evaluation \emph{end}. The task requires the model to perceive game state from pixels alone---no privileged API access---and produce a persistent, stateful controller under closed-loop execution. Synthesizing such a controller is not a one-shot code generation problem: implicit game parameters must be discovered empirically through interaction. PlayScriptBench therefore structures the task as a four-stage pipeline (Figure~\ref{fig:pipeline})---\emph{Explore}, \emph{Perception}, \emph{API Construction}, and \emph{Strategy Composition}---that decomposes controller synthesis into perception, interface construction, and strategic reasoning, each producing an independently scorable artifact. We further decouple visual grounding from strategic reasoning by scoring each dimension separately, enabling fine-grained failure diagnosis that a single end-to-end score collapses. A \emph{Joint score} (Perception $\times$ Strategy under Oracle perception) isolates strategy competence from perception failures, dropping to zero whenever either component fails.

We evaluate frontier VLMs on a diverse set of 16 games spanning 4 genres. Our results reveal a substantial gap between passive visual game understanding and executable closed-loop game control. Even strong VLMs struggle on our benchmark: models that fluently describe a static screenshot frequently fail to produce controllers that act reliably under live gameplay, suggesting that closed-loop debugging remains an open challenge.

Our contributions are as follows.
\begin{enumerate}
\item A four-stage \emph{Explore--Perception--API Construction--Strategy Composition} pipeline that enables VLMs to develop fully functional game controllers, with each stage producing a diagnosable artifact.
\item PlayScriptBench, a complementary benchmark on 16 deterministic, bounded-state HTML5 games spanning 4 genres, behind a uniform browser-based sandbox that pauses the game during inference so evaluation reflects what the agent decides, not how fast it responds.
\item An initial evaluation of 13 frontier VLMs exposing failure modes in perceptual grounding, strategic reasoning, and implicit parameter reasoning, with a Joint score (Perception $\times$ Strategy under Oracle perception) as a diagnostic measure of strategy competence.
\end{enumerate}

% =====================================================================
% Related Work
% =====================================================================
\section{Related Work}

\textbf{VLM and LLM gameplay agents.}
A growing body of work deploys LLMs and VLMs as gameplay agents. Voyager~\cite{wangVoyagerOpenEndedEmbodied2023}, JARVIS-1~\cite{wangJARVIS1OpenWorldMultitask2023a}, and MineDojo~\cite{fanMineDojoBuildingOpenEnded2022} use LLM planners in Minecraft but rely on privileged game state; Cradle~\cite{tanCradleEmpoweringFoundation2024} and STEVE-1~\cite{lifshitzSTEVE1GenerativeModel2024a} adopt a strict screenshot-in, action-out interface, demonstrating vision-only play is feasible. GameWorld~\cite{ouyangGameWorldStandardizedVerifiable2026}, VideoGameBench~\cite{zhangVideoGameBenchCanVisionLanguage2026}, Orak~\cite{parkOrakFoundationalBenchmark2026}, LMGame-Bench~\cite{huLmgameBenchHowGood2025}, GameDevBench~\cite{chiGameDevBenchEvaluatingAgentic2026}, and MCU~\cite{zhengMCUEvaluationFramework2023a} evaluate VLMs on large game suites under vision-only protocols, but their agents predict actions one step at a time. PlayScriptBench instead requires \emph{program synthesis} of a complete controller, forcing explicit perceptual grounding via OpenCV-based code, and separates parameter discovery from controller synthesis for finer-grained diagnosis. The two paradigms are complementary: per-frame inference excels at real-time reactivity, while script synthesis tests the ability to compress visual reasoning into persistent code.

\textbf{Visual program synthesis and code-synthesis evaluators.}
A parallel lineage uses LLMs to generate executable programs from visual inputs. ViperGPT~\cite{surisViperGPTVisualInference}, VisProg~\cite{guptaVisualProgrammingCompositional2022}, Chameleon~\cite{luChameleonPlugandPlayCompositional2023}, ProgPrompt~\cite{singhProgPromptGeneratingSituated2022}, and CodeVQA~\cite{subramanianModularVisualQuestion2023a} ground language in executable visual modules, but target \emph{stateless, single-inference} tasks. A related thread evaluates generated code by executing it in interactive environments: Voyager~\cite{wangVoyagerOpenEndedEmbodied2023} synthesizes skill libraries from privileged Minecraft state; Eureka~\cite{maEUREKAHUMANLEVELREWARD} synthesizes reward functions for physics simulations; Code World Models~\cite{lehrachCodeWorldModels2025} learn code-based world models; PlayCoder~\cite{pengPlayCoderMakingLLMGenerated2026} and ProxyWar~\cite{pengProxyWarDynamicAssessment2026} evaluate code by execution outcome in games and GUIs. PlayScriptBench extends both lines to \emph{stateful, closed-loop} control under a strict \emph{vision-only} constraint---no DOM, memory, or game-state access---structured as a four-stage pipeline producing independently scorable artifacts, exposing failure modes (drift accumulation, temporal grounding, strategic reasoning) absent from single-shot VPS or privileged-state evaluators.

\textbf{General agent benchmarks.}
WebArena~\cite{zhouWEBARREALISTIC2024a}, VisualWebArena~\cite{kohVisualWebArenaEvaluatingMultimodal2024}, OSWorld~\cite{xieOSWorldBenchmarkingMultimodal2024}, and AgentGym~\cite{xiAgentGymEvaluatingTraining2025} evaluate multimodal agents in web or desktop environments, typically granting access to DOM or accessibility trees alongside screenshots, which already encode much of the task-critical information and reduce the perceptual challenge. Text-based game benchmarks such as TextWorld~\cite{coteTextWorldLearningEnvironment2019} and ALFWorld~\cite{shridharALFWorldAligningText2021} take the opposite approach, stripping away visual input entirely. PlayScriptBench targets the underserved middle ground: visual environments where agents must extract game state entirely from pixels, without the symbolic abstraction of text games or the structured state access of web agents. GameplayQA~\cite{wangGameplayQABenchmarkingFramework2026} benchmarks decision-dense video understanding in 3D game environments in a passive QA format, whereas PlayScriptBench requires closed-loop executable scripts.

% =====================================================================
% Game Agent
% =====================================================================
\begin{figure*}[t]
  \centering
  \includegraphics[width=\textwidth]{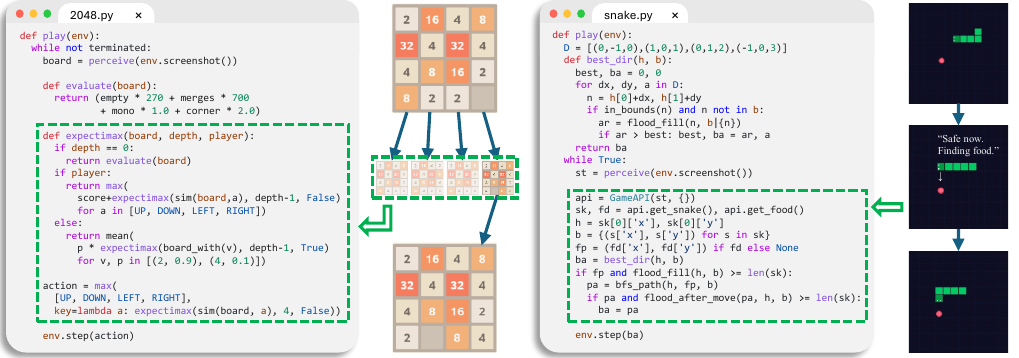}
  \caption{Stage~3 controller script and game sandbox interaction. Left: a 2048 controller generated by Claude Sonnet 4.6; right: a Snake controller generated by GPT-5.6 Luna.}
  \label{fig:stage3_strategy}
\end{figure*}

\section{Game Agent}

PlayScriptBench evaluates VLMs as game agents under a strict protocol. This section formalizes the agent artifacts, the evaluation harness, and the oracle--VLM probe mechanism that enables fine-grained failure attribution.

\subsection{Agent Artifacts}
\label{sec:artifacts}

Unlike GameWorld~\cite{ouyangGameWorldStandardizedVerifiable2026}, which studies both computer-use agents (CUA) that emit low-level keyboard/mouse controls and general-agent baselines that predict one action per frame, PlayScriptBench requires the VLM to author \emph{four distinct Python artifacts} across the pipeline, each consumed by the next stage.

\textbf{Stage~0 -- Explore.} Given only the game rules and a representative screenshot, the VLM freely plays the game for a short rollout, observing game mechanics, visual elements, and control responses. An analyst VLM then distills the screenshots and stdout into a structured markdown summary covering game phases, action effects, visual layout, and dynamics. This summary---\emph{exploration insights}---is consumed by Stage~1 as qualitative context for writing the perception routine.

\textbf{Stage~1 -- Perception.} Injected with the exploration insights from Stage~0, the VLM writes a \texttt{perceive(frame)} function that extracts structured game-state attributes from a raw RGB frame. The function receives two inputs beyond the frame itself: (i)~\emph{exploration insights} distilled by Stage~0, which provide qualitative understanding of game mechanics, visual layout, and action effects; and (ii)~\emph{sample frames} from a curated dataset, giving the VLM visual reference for colors, geometry, and anti-aliasing. The function must identify and localize all game-relevant elements---player position, obstacles, collectibles, UI indicators---and return a typed dataclass instance. For games requiring frame-to-frame deltas (e.g., motion-based detection), the signature extends to \texttt{perceive(frame, prev\_frame)}. A companion Constants task asks the VLM to fill in a \texttt{Constants} dataclass with concrete values for geometric quantities (positions, sizes) and physical rates (e.g., bridge extension speed), derived either from visual observation or from experiment samples embedded in the prompt.

\textbf{Stage~2 -- API Construction.} The VLM writes a \texttt{GameAPI} class that reshapes raw perceptual outputs into strategy-facing interfaces. For example, if Stage~1 detects ``block at position \texttt{(x, y)}'', Stage~2 might expose methods like \texttt{get\_nearest\_obstacle()} or \texttt{is\_jump\_safe()}. The class is instantiated on each frame's state (a \texttt{SimpleNamespace} derived from the perception output) and its methods are compared against an Oracle reference implementation.

\textbf{Stage~3 -- Strategy Composition.} The VLM writes a \texttt{play(env)} function that implements the game strategy. The controller is executed end-to-end in the sandbox; the harness captures score and survival time, and each frame is captured as an RGB array and passed to \texttt{perceive()} as input (Figure~\ref{fig:stage3_strategy}). The best controller is then executed across multiple seeds for robustness verification.

This decomposition makes failure attribution concrete: a model that perceives accurately but plays poorly has a strategy defect, while one that perceives poorly cannot produce a viable controller---a distinction that any single end-to-end score collapses.

\subsection{Evaluation Harness}
\label{sec:harness}

Foundation models alone are insufficient for sustained gameplay: the agent needs structured prompts and persistent memory to author coherent controllers. Therefore, we wrap each model in a shared evaluation harness that standardizes these components across all models.

\subsubsection{Structured Prompt}
To reduce prompt-induced variance across models and games, each stage uses a fixed prompt template that includes Game Rules, the stage-specific specification file (perception\_spec.py, api\_spec.py, or the combined prepend code), and Output Format. The template structure stays constant across all experiments; only the game-specific content is swapped per configuration, keeping cross-model comparisons controlled.

\subsubsection{Prepend Code Injection}
Unlike per-step agents that receive raw observations, PlayScriptBench's synthesized controllers operate on structured state. The harness injects a \emph{prepend code block} into the execution scope of each \texttt{play()} function, containing the perception specification, the chosen \texttt{perceive()} implementation (Oracle or VLM-authored), an adapter that wraps dict returns in \texttt{SimpleNamespace} for attribute access, and discovered constants. This mechanism decouples perception from strategy: the \texttt{play()} function calls \texttt{perceive(frame)} and operates on the returned state object without knowing whether the perception layer is Oracle-provided or VLM-authored.

\subsection{Oracle--VLM Probe Mechanism}
\label{sec:probe}

Without decoupling, the strategy score inherently conflates visual grounding with strategic reasoning, so a low score cannot distinguish between poor perception and poor strategic planning. A key design choice is the ability to \emph{swap the perception layer} in Stage~3, enabling two evaluation modes that isolate different failure modes. In the default oracle mode, Stage~3 receives the Oracle's \texttt{perceive()} and \texttt{GameAPI}, so the \texttt{play()} function operates on ground-truth state, measuring strategic reasoning in isolation. A \texttt{--use-vlm-probe} mode instead plugs in the VLM's own Stage~1 and Stage~2 artifacts, so errors propagate end-to-end and the full pipeline is tested. A controller must both perceive and act correctly to succeed; the \emph{Joint score} ($\mathrm{Obs.} \times \mathrm{Stra.}$, where $\mathrm{Obs.} = (\mathrm{Per.} + \mathrm{API.})/2$) collapses to zero whenever either component fails---a diagnostic measure of strategy competence under Oracle perception.

% =====================================================================
% The PlayScriptBench Benchmark
% =====================================================================
\section{PlayScriptBench Benchmark}

This section describes the benchmark design, the game suite, the browser-based sandbox environment, and the four-stage evaluation pipeline.

\subsection{Benchmark Design}
\label{sec:design}

PlayScriptBench evaluates a VLM's ability to author a closed-loop controller for an unfamiliar HTML5 game, given only the game's rules and a single representative screenshot. The benchmark enforces a strict protocol: the agent receives only raw RGB frames and a keyboard/mouse action interface, with no access to source code, runtime memory, or any state-querying API.

Three principles guide the design. First, perception must live inside the synthesized controller---unlike benchmarks that grant access to DOM trees, accessibility trees, or game-specific APIs, PlayScriptBench requires the model to implement its own OpenCV-style perception routines. Second, evaluation must be behavioral---controllers are graded by execution score through direct gameplay, not by code quality or textual output. Third, diagnosis must be fine-grained---the four-stage pipeline separates parameter discovery from controller construction, enabling attribution of failure to perception, reasoning, or control.

\subsection{Games Environment and Suite}
\label{sec:games}

The 16 games currently supported by PlayScriptBench span four genres: \emph{runner} (continuous scrolling with reflexive obstacle avoidance, e.g., Flappy Bird, Geometry Dash, Chrome Dino), \emph{arcade} (discrete turns or real-time scoring in a fixed arena, e.g., Pac-Man, Snake, Knife Hit), \emph{puzzle} (logical reasoning or optimization over a static or semi-static board, e.g., 2048, Minesweeper, Tetris), and \emph{platformer} (vertical or horizontal navigation across platforms with physics-driven movement, e.g., Doodle Jump, Helix Jump). Games are rendered by a mix of DOM, HTML5 Canvas, and Three.js engines, and accept either keyboard or mouse input, forcing the agent to adapt its perception and control strategy to the underlying representation.

Each game ships with a \emph{game configuration} that specifies the viewport, action mapping, and any game-specific status flags consumed by the evaluation harness.

\subsection{Browser-Based Sandbox Environment}
\label{sec:sandbox}

The central design goal of the sandbox is to decouple agent decision quality from inference speed. In real-time games, a slower model faces a harder game state by the time it acts, conflating thinking time with gameplay ability. To eliminate this confound, the sandbox pauses game execution during controller synthesis, so every agent faces identical game dynamics regardless of response latency. Scores then reflect what the agent decides, not how fast it responds.

\subsection{Evaluation Pipeline}
\label{sec:pipeline}

The benchmark decomposes evaluation into four stages, each producing an artifact that the next stage consumes. All four stages are executed by the same VLM, with no separate models assigned to individual stages. This decomposition isolates whether failure stems from perceptual grounding, API construction, or strategic reasoning---diagnoses a single end-to-end score cannot provide. The pipeline supports two evaluation modes: \emph{oracle mode} (default) tests strategic reasoning alone by providing Oracle perception, while \emph{VLM probe mode} tests the full pipeline end-to-end by using the VLM's own perception artifacts.

\subsubsection{Stage 0 -- Explore}
The analyst VLM distills the rollout screenshots and stdout into a structured markdown summary. The summary is evaluated for coverage (game phases, action effects, visual layout, dynamics) and accuracy against the game specification. The resulting \emph{exploration insights} artifact is consumed by Stage~1 as qualitative context.

\subsubsection{Stage 1 -- Perception}
The VLM produces a \texttt{perceive()} function and a \texttt{Constants} dataclass. The grader evaluates perception accuracy by running the VLM-authored function on curated frames with known ground truth, measuring per-attribute precision and recall against specification tolerances. The Constants task is scored by comparing filled values against verified reference constants.

\subsubsection{Stage 2 -- API Construction}
The VLM produces a \texttt{GameAPI} class that exposes strategy-facing interfaces over perceptual outputs. The grader instantiates the VLM-authored API on each frame's oracle perception output, invokes every method, and compares results against the oracle reference implementation. Per-method functional correctness is measured, aggregating into an overall API score.

\subsubsection{Stage 3 -- Strategy Composition}
The VLM produces a \texttt{play(env)} controller. Evaluation proceeds in two phases. Phase~1 (Generation): The controller is executed end-to-end in the sandbox; the harness captures score and survival time, and each frame is passed to \texttt{perceive()} as input. Phase~2 (Multi-Run Verification): The controller is executed $N$ times on a seed set, reporting score and survival time as mean$\pm$standard deviation along with a discrete execution status (pass, run-time error, timeout). Reporting variance is essential: many HTML5 games are stochastic, and a controller that occasionally scores well by luck is meaningfully different from one that scores reliably.

A key design choice is the ability to swap the perception layer in Stage~3. By default (oracle mode), the controller receives the Oracle's \texttt{perceive()} and \texttt{GameAPI}, testing strategic reasoning alone. With the \texttt{--use-vlm-probe} flag, the controller receives the VLM's own Stage~1 and Stage~2 artifacts, testing the full pipeline end-to-end. Comparing scores between these two modes reveals whether failure stems from perceptual grounding or strategic reasoning.

% =====================================================================
% Experiments
% =====================================================================
\section{Experiments}

We evaluate thirteen frontier VLMs from multiple providers on the PlayScriptBench game suite, measuring their ability to author self-contained controllers: GPT-5.6 Luna and GPT-5.4 Mini~\cite{singhOpenAIGPT5System2026}, Claude Sonnet 4.6 and Claude Sonnet 5~\cite{ClaudeSonnet}, Doubao Seed 2.1 Turbo~\cite{ByteDanceSeed}, Qwen 3.6 Plus~\cite{Qwen36LLMTechnical} and Qwen 3.5 Flash~\cite{teamQwenStudio2026}, Gemini 3.1 Flash Lite and Gemini 3.5 Flash~\cite{teamGeminiFamilyHighly2025}, GLM-4.6V~\cite{teamGLM45VGLM41VThinkingVersatile2026}, ERNIE 5.0~\cite{wangERNIE50Technical2026}, Kimi K2.5~\cite{teamKimiK25Visual2026}, and MiniMax-M3~\cite{minimaxMiniMaxM3Frontier}. Each model's final script is evaluated over 10 seeds; the mean score across seeds serves as the Stage~3 strategy score.

\subsection{Main Results}
\label{sec:main}

\begin{table*}[t]
\centering
\setlength{\tabcolsep}{2.5pt}
\small
\begin{tabular}{lcc@{\hskip 0.8em}cc@{\hskip 0.8em}cc@{\hskip 0.8em}cc@{\hskip 0.8em}ccc}
  \toprule
  & \multicolumn{2}{c}{\textbf{Runner}} & \multicolumn{2}{c}{\textbf{Arcade}} & \multicolumn{2}{c}{\textbf{Puzzle}} & \multicolumn{2}{c}{\textbf{Platformer}} & \multicolumn{3}{c}{\textbf{Overall}} \\
  \cmidrule(lr){2-3} \cmidrule(lr){4-5} \cmidrule(lr){6-7} \cmidrule(lr){8-9} \cmidrule(lr){10-12}
  \textbf{Model} & \textbf{Obs.} & \textbf{Stra.} & \textbf{Obs.} & \textbf{Stra.} & \textbf{Obs.} & \textbf{Stra.} & \textbf{Obs.} & \textbf{Stra.} & \textbf{Obs.} & \textbf{Stra.} & \textbf{Joint} \\
  \midrule
  MiniMax-M3                & \underline{0.65} & 0.148 & 0.54 & 0.334 & 0.40 & 0.493 & 0.48 & 0.231 & 0.52 & 0.302 & 0.157 \\
  Claude Sonnet 4.6         & 0.59 & 0.185 & \textbf{0.65} & 0.435 & 0.43 & \underline{0.648} & 0.44 & 0.390 & \underline{0.53} & 0.414 & 0.218 \\
  Claude Sonnet 5           & 0.48 & 0.093 & 0.56 & 0.139 & 0.33 & 0.282 & 0.45 & 0.394 & 0.46 & 0.227 & 0.103 \\
  Doubao Seed 2.1 Turbo      & 0.56 & 0.404 & 0.59 & 0.331 & \underline{0.52} & 0.590 & 0.39 & 0.437 & 0.51 & \underline{0.441} & \underline{0.227} \\
  ERNIE 5.0                 & 0.49 & 0.333 & 0.42 & 0.308 & 0.42 & 0.270 & 0.48 & \textbf{0.506} & 0.45 & 0.354 & 0.160 \\
  Gemini 3.1 Flash Lite     & 0.57 & 0.074 & 0.44 & 0.207 & 0.26 & 0.207 & 0.30 & 0.485 & 0.39 & 0.243 & 0.095 \\
  Gemini 3.5 Flash          & 0.24 & \underline{0.639} & 0.41 & \underline{0.495} & 0.51 & 0.144 & 0.22 & 0.161 & 0.34 & 0.360 & 0.123 \\
  GLM-4.6V                  & 0.47 & 0.185 & 0.38 & 0.096 & 0.18 & 0.000 & 0.40 & 0.297 & 0.36 & 0.144 & 0.051 \\
  GPT-5.4 Mini              & 0.56 & 0.130 & \underline{0.62} & 0.326 & 0.25 & 0.590 & \textbf{0.61} & 0.325 & 0.51 & 0.343 & 0.174 \\
  GPT-5.6 Luna              & 0.60 & 0.204 & 0.61 & 0.159 & 0.44 & 0.306 & 0.46 & 0.349 & \textbf{0.53} & 0.255 & 0.134 \\
  Kimi K2.5                 & 0.60 & 0.333 & 0.55 & 0.390 & \textbf{0.54} & \textbf{0.658} & 0.35 & 0.358 & 0.51 & 0.435 & 0.223 \\
  Qwen 3.5 Flash            & \textbf{0.67} & 0.148 & 0.49 & 0.276 & 0.20 & 0.197 & 0.38 & \underline{0.486} & 0.44 & 0.277 & 0.120 \\
  Qwen 3.6 Plus             & 0.52 & \textbf{0.667} & 0.55 & \textbf{0.665} & 0.31 & 0.508 & \underline{0.50} & 0.408 & 0.47 & \textbf{0.562} & \textbf{0.265} \\
  \midrule
  Avg.                 & 0.54 & 0.273 & 0.52 & 0.320 & 0.37 & 0.376 & 0.42 & 0.371 & 0.46 & 0.335 & 0.155 \\
  \bottomrule
\end{tabular}
\caption{
  Stage-3 evaluation results across the 16-game suite.
  \textbf{Obs.}~(Observation): end-to-end state extraction quality, averaging perception and API accuracy.
  \textbf{Stra.}~(Strategy): min-max normalized game score across 10 random seeds, using Oracle perception; bounds are the min and max raw game scores among all evaluated models on that game.
  \textbf{Joint}: a composite diagnostic score, $\frac{1}{G}\sum_{g}\mathrm{Obs}_g \times \mathrm{Stra}_g$, where Stra.\ is measured under Oracle perception.
  \textbf{Bold} indicates first place; \underline{underline} indicates second place per column.
}
\label{tab:main_results}
\end{table*}

Table~\ref{tab:main_results} shows Stage~3 strategy scores across the 16-game suite. We report three metrics at the observation level: \textbf{Per.} (Perception accuracy), the precision and recall of the VLM-authored \texttt{perceive()} function against ground-truth game-state attributes; \textbf{API.} (API correctness), the fraction of \texttt{GameAPI} methods that produce correct outputs when instantiated on oracle perception; and \textbf{Obs.} (Observation score), the arithmetic mean of Per. and API. for each game, $\mathrm{Obs}_g = (\mathrm{Per}_g + \mathrm{API}_g)/2$, measuring end-to-end state extraction quality. We further report \textbf{Joint}, the mean of per-game (Observation $\times$ Strategy) products, $\frac{1}{G}\sum_{g} \mathrm{Obs}_g \times \mathrm{Stra}_g$, as a diagnostic measure of strategy competence under Oracle perception. The models show complementary strengths across genres.

Several patterns emerge from the results. Most notably, observation capability does not translate linearly into final strategy performance. GPT-5.6 Luna achieves the highest overall Obs.\ (0.53, tied with Claude Sonnet 4.6) yet ranks only tenth in Stra.\ (0.255) and eighth in joint score (0.134), while Qwen 3.6 Plus attains merely mid-tier Obs.\ (0.47) but dominates Stra.\ in Runner and Arcade, yielding the top joint score (0.265). This dissociation indicates that the benchmark's bottleneck is not visual recognition in itself, but the ability to convert partial perceptual state into effective sequential decisions, as further confirmed by the pipeline ablation in Table~\ref{tab:pipeline_ablation}. Genre also matters: Puzzles and Platformers achieve the highest average Stra.\ (0.376 and 0.371), while Puzzles show the lowest average Obs.\ (0.37), reflecting the role of bounded state spaces in making deterministic puzzle solvers tractable. Finally, a striking failure mode appears in models such as GLM-4.6V, which maintain non-trivial Obs.\ (0.36) yet produce near-zero Stra.\ (0.144)---demonstrating that passing the observation stage is necessary but far from sufficient for downstream gameplay. Overall, the average joint score across all models stands at only 0.155, underscoring that current frontier VLMs remain far from mastering even simple game environments end-to-end.

% =====================================================================
% Ablation Studies
% =====================================================================
\section{Ablation Studies}
\label{sec:ablation}

We ablate each stage of the four-stage pipeline to quantify its contribution to end-to-end performance. Table~\ref{tab:pipeline_ablation} summarizes the results on GPT-5.6 Luna and Qwen 3.6 Plus (16-game average). The ``Full Pipeline'' row uses Oracle perception; its per-game Per.\ and API.\ are averaged directly over 16 games (see Table~\ref{tab:pipeline_ablation} caption), so they are not directly comparable to the genre-averaged Overall Obs.\ in Table~\ref{tab:main_results}. The ``w/o Oracle'' row replaces Oracle with the VLM's own Stage~1/2 artifacts, revealing the perception bottleneck.

\begin{table}[t]
\centering
\setlength{\tabcolsep}{3pt}
\small
\begin{tabular}{lccc ccc@{\hskip 1em}ccc ccc}
  \toprule
   & \multicolumn{6}{c}{\textbf{GPT-5.6 Luna}} & \multicolumn{6}{c}{\textbf{Qwen 3.6 Plus}} \\
  \cmidrule(lr){2-7} \cmidrule(lr){8-13}
  \textbf{Config} & \textbf{Per.} & \textbf{API.} & \textbf{Stra.} & $\downarrow$ & $\rightarrow$ & $\uparrow$ & \textbf{Per.} & \textbf{API.} & \textbf{Stra.} & $\downarrow$ & $\rightarrow$ & $\uparrow$ \\
  \midrule
  Full Pipeline   & 0.793 & 0.576 & 0.255 & --- & --- & --- & 0.609 & 0.334 & 0.600 & --- & --- & --- \\
  w/o Stage 0     & 0.463 & 0.327 & ---   & 6   & 9   & 1   & 0.458 & 0.231 & ---   & 8   & 7   & 1   \\
  w/o Stage 1     & ---   & 0.251 & ---   & 6   & 10  & 0   & ---   & 0.149 & ---   & 10  & 6   & 0   \\
  w/o Stage 2     & ---   & ---   & 0.036 & 7   & 9   & 0   & ---   & ---   & 0.120 & 9   & 6   & 1   \\
  w/o Oracle      & ---   & ---   & 0.013 & 13  & 3   & 0   & ---   & ---   & 0.052 & 14  & 2   & 0   \\
  \bottomrule
\end{tabular}
\caption{
  Pipeline ablation on GPT-5.6 Luna (best perception) and Qwen 3.6 Plus (best strategy), 16-game average. \textbf{w/o Stage 0/1/2}: removes Explore/Perception/API Construction; \textbf{w/o Oracle}: replaces Oracle perception with the VLM's own Stage~1/2 artifacts. \textbf{Per.}/\textbf{API.}/\textbf{Stra.}: perception accuracy (Stage~1), API correctness (Stage~2), and min-max normalized strategy score (Stage~3, Oracle perception). Per.\ and API.\ are averaged directly across 16 games, so they differ from the genre-averaged Overall Obs.\ in Table~\ref{tab:main_results}. Arrows ($\downarrow$, $\rightarrow$, $\uparrow$): among games with valid scores, the number where score decreases ($>10\%$)/stays ($\pm10\%$)/increases ($>10\%$) vs.\ Full Pipeline.
}
\label{tab:pipeline_ablation}
\end{table}

\subsection{Explore Stage (Stage~0)}
\label{sec:explore_ablation}

Without Explore, Stage~1 receives only the game rules and perception spec, forcing the model to infer visual properties (color signatures, object positions) solely from the spec's dataclass definitions. With Explore, the model additionally receives structured observations---e.g., color signatures and spatial layouts of game objects---that provide qualitative grounding for the perception routine.

The Explore benefit is most pronounced on games whose visual elements cannot be inferred from the spec alone, while games with visually self-documenting specs show smaller drops. On GPT-5.6 Luna, removing Explore reduces Obs.\ from 0.685 to 0.395 ($-42.3\%$), with Per.\ dropping from 0.793 to 0.463 and API.\ from 0.576 to 0.327. On Qwen 3.6 Plus, Obs.\ drops from 0.472 to 0.345 ($-26.9\%$), a more modest decline consistent with this model's stronger prior for visual property extraction from text. The arrow statistics in Table~\ref{tab:pipeline_ablation} corroborate this: on GPT-5.6 Luna, 6 out of 16 games degrade by more than 10\% without Explore while only 1 improves; on Qwen 3.6 Plus, 8 degrade and 1 improves.

\subsection{Perception and API Construction (Stages~1--2)}
\label{sec:perception_ablation}

Two patterns emerge from the perception and API construction stages. First, Stage~2 accuracy drops sharply when Stage~1 is imperfect: on Knife Hit, 74\% perception accuracy yields only 3\% API accuracy, indicating that perception errors are amplified rather than corrected by the reasoning layer. Second, even the sole perfect game (Match-3 at 100\% across both stages) leaves the strategy layer to handle residual execution challenges---the gap between perfect observation and high gameplay score remains non-trivial.

Removing perception entirely (w/o Stage~1) collapses API accuracy to 0.251/0.149 for GPT-5.6 Luna/Qwen 3.6 Plus since the VLM cannot provide perceptual state to the API layer. Removing GameAPI (w/o Stage~2) destroys Stra.\ (0.255$\to$0.036, $-86\%$ for GPT-5.6 Luna) as the model cannot reason over raw state tuples.

The Probe stage provides measurable benefits when visual parameters are discoverable. For games with clear color signatures and grid geometry (Tetris, Match-3), injected measurements improve perception accuracy. However, for games with dynamic visual elements (Flappy Bird, Knife Hit), the Probe stage contributes less, as the critical parameters change during gameplay.

\subsection{Oracle vs.\ VLM Probe}
\label{sec:oracle_ablation}

Replacing the Oracle perception with the VLM's own Stage~1 and Stage~2 artifacts (w/o Oracle) collapses strategy score to near zero (0.013 for GPT-5.6 Luna, 0.052 for Qwen 3.6 Plus), with the majority of games showing degradation. This confirms that the strategy layer's performance is tightly coupled to perception quality: even models with strong perception (Stage~1 accuracy 0.793) produce controllers that fail when operating over their own perceptual outputs. The gap between Oracle mode (Stra.\ 0.255) and VLM probe mode (Stra.\ 0.013) for GPT-5.6 Luna indicates that perception errors propagate through API construction into catastrophic strategy failures, validating the pipeline's decomposition of these concerns.

\subsection{Resource Efficiency vs.\ VLM in-loop}
\label{sec:resource_ablation}

\begin{table}[t]
\centering
\small
\setlength{\tabcolsep}{5pt}
\begin{tabular}{lccc@{\hskip 1.5em}cccc}
  \toprule
   & \multicolumn{3}{c}{\textbf{1 Seed}} & \multicolumn{4}{c}{\textbf{10 Seeds}} \\
  \cmidrule(lr){2-4} \cmidrule(lr){5-8}
  \textbf{Method} & \textbf{Calls} & \textbf{Tokens} & \textbf{Time} & \textbf{Calls} & \textbf{Tokens} & \textbf{Time} & \textbf{Stra.} \\
   &  & (K) & (min) &  & (K) & (min) &  \\
  \midrule
  Our Method    & 5.0   & 27.4    & 4.4  & 5.0    & 27.4    & 19.7 & 0.600 \\
  VLM in-loop   & 214.1 & 2270.6  & 4.2  & 2141.3 & 22706.3 & 42.1 & 0.357 \\
  \bottomrule
\end{tabular}
\caption{
  Resource consumption on Qwen 3.6 Plus (16-game average). For VLM in-loop, the 10-seed Tokens value is cumulative across all seeds. Stra.: min-max normalized strategy score (10-seed mode only).
}
\label{tab:resource_ablation}
\end{table}

Table~\ref{tab:resource_ablation} compares our pipeline against a VLM in-loop baseline on Qwen 3.6 Plus, which queries the VLM for visual observations at 0.1~s intervals throughout gameplay. With a single seed, our approach reduces VLM calls by 42.8$\times$ (5.0 vs.\ 214.1) and tokens by 82.9$\times$ (27.4K vs.\ 2270.6K), though with modest wall-clock overhead (4.4 vs.\ 4.2 min). The key advantage emerges at 10 seeds: because our controller is reusable, it requires only the same 5.0 VLM calls per game, while the in-loop baseline re-queries at every step for each seed (2141.3 calls). This yields a 2.14$\times$ speedup (19.7 vs.\ 42.1 min) alongside a 68.1\% higher strategy score (0.600 vs.\ 0.357). Note that our Stra.\ is measured under Oracle perception while the in-loop baseline uses its own perception, so this gap reflects amortization under non-matched state access.

% =====================================================================
% Conclusion
% =====================================================================
\section{Conclusions and Limitations}

PlayScriptBench introduces a \emph{complementary} script-based evaluation protocol for VLM gameplay agents, shifting from per-frame decision-making to reusable controller synthesis on the class of deterministic, bounded-state games where static controllers can play reliably. Our four-stage pipeline---Explore, Perception, API Construction, and Strategy Composition---decomposes controller synthesis into perception, interface construction, and strategic reasoning, enabling fine-grained diagnosis of failure modes that a single end-to-end score cannot reveal. Our evaluation of frontier VLMs on 16 HTML5 games reveals a substantial gap between passive image understanding and active closed-loop control: perceptual anchoring is fragile, and perception accuracy is necessary but not sufficient, with the bottleneck lying in API construction and strategic reasoning.

PlayScriptBench targets deterministic, bounded-state games, and the synthesized controllers cannot adapt their strategy online. Future work includes introducing online VLM feedback for runtime script refinement and scaling the game suite to cover more genres.

% =====================================================================
% AI use statement (kept from the ICLR submission)
% =====================================================================
\section*{AI Use Statement}
In this work, we used generative AI tools for the following tasks with \emph{required} disclosure: \emph{implementing methods}---the four-stage pipeline artifacts (perception routines, GameAPI implementations, and strategy controllers) evaluated in our benchmark are themselves VLM-authored code, which is the object of study rather than a tool used to conduct the study; and \emph{interpreting results}---we consulted LLMs when drafting prose that describes experimental patterns and failure modes. We have not used generative AI tools for generating synthetic datasets, formulating theoretical models or mathematical claims, or assisting with proofs, and these are not applicable to this work. Additionally, we used generative AI tools for the following tasks with \emph{recommended} disclosure: \emph{drafting parts of the research paper}, \emph{editing to improve readability}, \emph{formatting references}, and \emph{creating or modifying scientific figures}. We have reviewed all AI-assisted work: all reported numbers were regenerated by the authors from the evaluation harness and cross-checked against the raw logs, all VLM-authored controller code was executed and graded by the authors' own harness (not by an LLM), and every claim in the paper was verified by manual inspection of the underlying artifacts. We take responsibility for the final content of this work, including text, claims, or artifacts produced with the aid of generative AI.

% =====================================================================
% References
% =====================================================================
\bibliography{ScriptBench}

% =====================================================================
% Appendix
% =====================================================================
\clearpage
\appendix
\section{Supplementary Material}
\subsection{Game Suite}
\label{sec:appendix_game_suite}

Table~\ref{tab:games} lists the 16 games supported by PlayScriptBench, spanning four genres: \emph{runner}, \emph{arcade}, \emph{puzzle}, and \emph{platformer}. Games are rendered by a mix of DOM, HTML5 Canvas, and Three.js engines, and accept either keyboard or mouse input, forcing the agent to adapt its perception and control strategy to the underlying representation.

\begin{table}[!h]
\centering
\small
\begin{tabular}{llll}
  \toprule
  \textbf{Game} & \textbf{Genre} & \textbf{Engine} & \textbf{Input} \\
  \midrule
  Flappy Bird      & Runner         & Canvas         & Keyboard       \\
  2048             & Puzzle         & DOM + CSS      & Keyboard       \\
  Pac-Man          & Arcade         & DOM grid       & Keyboard       \\
  Snake            & Arcade         & Canvas         & Keyboard       \\
  Tetris           & Puzzle         & Canvas         & Keyboard       \\
  Minesweeper      & Puzzle         & DOM grid       & Mouse          \\
  Doodle Jump      & Platformer     & Canvas         & Keyboard       \\
  Stack2D          & Platformer     & Canvas         & Keyboard       \\
  Stick Hero       & Platformer     & Canvas         & Mouse          \\
  Knife Hit        & Arcade         & Canvas         & Keyboard       \\
  Color Switch     & Runner         & Canvas         & Keyboard       \\
  Space Invaders   & Arcade         & Canvas         & Keyboard       \\
  Helix Jump       & Platformer     & Three.js       & Mouse   \\
  Geometry Dash    & Runner         & Canvas         & Keyboard       \\
  Match-3         & Puzzle         & DOM grid       & Mouse          \\
  Chrome Dino      & Runner         & DOM + Canvas   & Keyboard       \\
  \bottomrule
\end{tabular}
\caption{PlayScriptBench game suite.}
\label{tab:games}
\end{table}

\subsection{Per-Game Stage~3 Strategy Scores}

Tables~\ref{tab:stage3_puzzle}, \ref{tab:stage3_platformer}, \ref{tab:stage3_runner}, and~\ref{tab:stage3_arcade} report the per-game Stage~3 strategy scores (mean and standard deviation over 10 screening seeds) for all models. These are the raw game scores used to compute the normalized strategy scores in the main results. The standard deviation quantifies the score variance across different random seeds, revealing that some games (e.g., 2048, Doodle Jump) exhibit high seed sensitivity while others (e.g., Chrome Dino, Minesweeper) are more deterministic.

\begin{table*}[tb]
\centering
\small
\setlength{\tabcolsep}{4pt}
\begin{tabular}{lcccc}
\toprule
\textbf{Model} & Chrome\_Dino & Color\_Switch & Flappy\_Bird & Geometry\_Dash \\
\midrule
MiniMax-M3 & 45.0 & 0.3 ($\pm$ 0.5) & 0 & 13.3 ($\pm$ 2.4) \\
Claude Sonnet 4.6 & 47.7 ($\pm$ 3.8) & 0 & 0 & 10.0 \\
Claude Sonnet 5 & 46.0 & 0 & 0 & 10.0 \\
Doubao Seed 2.1 Turbo & 45.0 & 0.7 ($\pm$ 0.5) & 0.6 ($\pm$ 0.9) & 15.0 \\
ERNIE 5.0 & 50.3 ($\pm$ 3.8) & 1.3 ($\pm$ 0.9) & 0 & 10.0 \\
Gemini 3.1 Flash Lite & 44.7 ($\pm$ 0.5) & 0 & 0 & 11.7 ($\pm$ 2.4) \\
Gemini 3.5 Flash & 46.0 & 1.0 ($\pm$ 0.8) & 1.0 & \textemdash \\
GLM-4.6V & 45.7 ($\pm$ 0.5) & 0 & 0.3 ($\pm$ 0.5) & 10.0 \\
GPT-5.4 Mini & 44.7 ($\pm$ 0.5) & 0.3 ($\pm$ 0.5) & 0 & 13.3 ($\pm$ 2.4) \\
GPT-5.6 Luna & 46.0 & 0 & 0.3 ($\pm$ 0.5) & 10.0 \\
Kimi K2.5 & 44.3 ($\pm$ 0.5) & 0.7 ($\pm$ 0.9) & 0 & 20.0 ($\pm$ 7.1) \\
Qwen 3.5 Flash & 46.0 & 0.3 ($\pm$ 0.5) & 0 & 11.7 ($\pm$ 2.4) \\
Qwen 3.6 Plus & 50.3 ($\pm$ 4.5) & 1.0 ($\pm$ 0.8) & 1.0 ($\pm$ 0.8) & 10.0 \\
\midrule
\textbf{Avg.} & 46.2 & 0.4 & 0.3 & 12.1 \\
\bottomrule
\end{tabular}
\caption{Per-game Stage~3 strategy scores on Runner games (mean $\pm$ std over 10 screening seeds). Dashes indicate games with no completed run.}
\label{tab:stage3_runner}
\end{table*}

\begin{table*}[tb]
\centering
\small
\setlength{\tabcolsep}{4pt}
\begin{tabular}{lcccc}
\toprule
\textbf{Model} & Pac\_Man & Snake & Knife\_Hit & Space\_Inv. \\
\midrule
MiniMax-M3 & 20.0 & 0 & 12.0 ($\pm$ 3.6) & 20.0 ($\pm$ 28.3) \\
Claude Sonnet 4.6 & 36.3 ($\pm$ 5.2) & 23.3 ($\pm$ 26.2) & 5.0 ($\pm$ 1.4) & 20.0 ($\pm$ 28.3) \\
Claude Sonnet 5 & \textemdash & \textemdash & 3.3 ($\pm$ 1.2) & \textemdash \\
Doubao Seed 2.1 Turbo & 0 & 0 & 4.7 ($\pm$ 1.2) & 61.7 ($\pm$ 49.4) \\
ERNIE 5.0 & 0 & 0 & 7.7 ($\pm$ 0.5) & 0 \\
Gemini 3.1 Flash Lite & 1.0 & 3.3 ($\pm$ 4.7) & 11.0 & 0 \\
Gemini 3.5 Flash & 31.3 ($\pm$ 0.5) & 0 & \textemdash & 20.0 ($\pm$ 28.3) \\
GLM-4.6V & 0 & 0 & 0 & 80.0 ($\pm$ 28.3) \\
GPT-5.4 Mini & 34.0 ($\pm$ 2.4) & 0 & 3.3 ($\pm$ 2.1) & 70.0 ($\pm$ 37.4) \\
GPT-5.6 Luna & 6.3 ($\pm$ 2.4) & 0 & 0 & 104.0 ($\pm$ 24.1) \\
Kimi K2.5 & 6.3 ($\pm$ 2.4) & 36.7 ($\pm$ 17.0) & 5.0 ($\pm$ 1.4) & 60.0 ($\pm$ 49.0) \\
Qwen 3.5 Flash & 10.3 ($\pm$ 6.6) & 0 & 0 & 100.0 ($\pm$ 28.3) \\
Qwen 3.6 Plus & 30.3 ($\pm$ 0.5) & 3.3 ($\pm$ 4.7) & 6.0 ($\pm$ 1.6) & 167.3 ($\pm$ 102.8) \\
\midrule
\textbf{Avg.} & 14.4 & 5.1 & 4.8 & 58.6 \\
\bottomrule
\end{tabular}
\caption{Per-game Stage~3 strategy scores on Arcade games (mean $\pm$ std over 10 screening seeds). Dashes indicate games with no completed run.}
\label{tab:stage3_arcade}
\end{table*}

\begin{table*}[tb]
\centering
\small
\setlength{\tabcolsep}{4pt}
\begin{tabular}{lcccc}
\toprule
\textbf{Model} & Match\_3 & Minesweeper & Tetris & 2048 \\
\midrule
MiniMax-M3 & 0 & 23.7 ($\pm$ 15.0) & 1950.0 ($\pm$ 334.2) & 1830.7 ($\pm$ 768.7) \\
Claude Sonnet 4.6 & 0 & 25.7 ($\pm$ 14.4) & 1950.0 ($\pm$ 334.2) & 4222.7 ($\pm$ 542.0) \\
Claude Sonnet 5 & 3613.3 ($\pm$ 1237.8) & \textemdash & \textemdash & \textemdash \\
Doubao Seed 2.1 Turbo & 12800.0 ($\pm$ 2490.4) & 25.0 ($\pm$ 16.1) & 0 & 2813.3 ($\pm$ 117.9) \\
ERNIE 5.0 & \textemdash & 0 & 683.3 ($\pm$ 458.9) & 2118.7 ($\pm$ 1351.3) \\
Gemini 3.1 Flash Lite & 2746.7 ($\pm$ 556.5) & 15.3 ($\pm$ 18.2) & 0 & 797.3 ($\pm$ 90.8) \\
Gemini 3.5 Flash & 5540.0 ($\pm$ 1206.1) & \textemdash & 0 & 0 \\
GLM-4.6V & 0 & 0 & 0 & 0 \\
GPT-5.4 Mini & 0 & 33.0 ($\pm$ 5.9) & 1883.3 ($\pm$ 246.1) & 2509.3 ($\pm$ 563.8) \\
GPT-5.6 Luna & 0 & 36.0 & 500.0 & 0 \\
Kimi K2.5 & 5213.3 ($\pm$ 1395.4) & 20.0 ($\pm$ 13.6) & 2216.7 ($\pm$ 143.4) & 2824.0 ($\pm$ 87.9) \\
Qwen 3.5 Flash & 0 & 1.0 & 0 & 3202.7 ($\pm$ 693.9) \\
Qwen 3.6 Plus & 0 & 23.0 ($\pm$ 14.4) & 2216.7 ($\pm$ 143.4) & 1664.0 ($\pm$ 319.9) \\
\midrule
\textbf{Avg.} & 2301.0 & 16.9 & 950.0 & 1798.8 \\
\bottomrule
\end{tabular}
\caption{Per-game Stage~3 strategy scores on Puzzle games (mean $\pm$ std over 10 screening seeds). Dashes indicate games with no completed run.}
\label{tab:stage3_puzzle}
\end{table*}

\begin{table*}[tb]
\centering
\small
\setlength{\tabcolsep}{4pt}
\begin{tabular}{lcccc}
\toprule
\textbf{Model} & Doodle\_Jump & Helix\_Jump & Stack2D & Stick\_Hero \\
\midrule
MiniMax-M3 & 47.3 ($\pm$ 23.6) & 2.0 ($\pm$ 2.2) & 5.3 ($\pm$ 1.9) & 0 \\
Claude Sonnet 4.6 & 379.0 ($\pm$ 479.9) & 0.3 ($\pm$ 0.5) & 5.3 ($\pm$ 2.5) & 3.0 ($\pm$ 2.9) \\
Claude Sonnet 5 & 269.7 ($\pm$ 187.3) & 1.7 ($\pm$ 1.9) & \textemdash & \textemdash \\
Doubao Seed 2.1 Turbo & 674.0 ($\pm$ 686.8) & 0.7 ($\pm$ 1.2) & 4.0 ($\pm$ 0.8) & 0 \\
ERNIE 5.0 & 202.0 ($\pm$ 88.7) & 2.3 ($\pm$ 3.3) & 2.3 ($\pm$ 0.5) & 9.0 ($\pm$ 9.2) \\
Gemini 3.1 Flash Lite & 762.7 ($\pm$ 841.0) & 1.0 ($\pm$ 1.4) & 2.3 ($\pm$ 0.9) & 4.0 ($\pm$ 3.7) \\
Gemini 3.5 Flash & 234.3 ($\pm$ 149.0) & 0 & 2.7 ($\pm$ 0.5) & 0 \\
GLM-4.6V & 181.7 ($\pm$ 134.7) & 2.3 ($\pm$ 3.3) & 0 & 0 \\
GPT-5.4 Mini & 478.7 ($\pm$ 225.7) & 1.3 ($\pm$ 1.6) & 1.0 & 5.0 ($\pm$ 3.6) \\
GPT-5.6 Luna & 634.0 ($\pm$ 513.8) & 0 & 3.0 & 1.3 ($\pm$ 0.9) \\
Kimi K2.5 & 549.7 ($\pm$ 233.7) & 1.3 ($\pm$ 1.9) & 2.0 ($\pm$ 0.8) & 2.3 ($\pm$ 2.1) \\
Qwen 3.5 Flash & 721.3 ($\pm$ 685.0) & 0.3 ($\pm$ 0.5) & 7.0 ($\pm$ 2.2) & 0 \\
Qwen 3.6 Plus & 269.7 ($\pm$ 210.3) & 2.7 ($\pm$ 2.9) & 3.3 ($\pm$ 0.5) & 3.3 ($\pm$ 2.4) \\
\midrule
\textbf{Avg.} & 415.7 & 1.0 & 3.2 & 2.2 \\
\bottomrule
\end{tabular}
\caption{Per-game Stage~3 strategy scores on Platformer games (mean $\pm$ std over 10 screening seeds). Dashes indicate games with no completed run.}
\label{tab:stage3_platformer}
\end{table*}

\subsection{Scoring Methodology}
\label{sec:scoring}

This appendix details the grading criteria for Per. (Stage~1), API. (Stage~2), and their aggregation into the composite metrics reported in the main results.

\subsubsection{Perception Score}

Per. measures how accurately the VLM-authored \texttt{perceive()} function extracts structured game state from raw screenshots. For each game, a curated set of frames with manually verified ground truth is used. The grader runs the VLM's \texttt{perceive()} on each frame and compares every declared attribute against the oracle, counting individual \emph{hits} and \emph{misses}. The per-game score is the ratio of total hits to total expected attributes across all frames:
\[
\mathrm{Per}_g = \frac{\text{total hits}}{\text{total expected attributes}}.
\]

\medskip
\noindent\textbf{Attribute types and comparison rules.}
Each attribute is compared according to its declared type:

\begin{itemize}
  \item \textbf{Numeric} (\texttt{int}, \texttt{float}): A hit is scored when $|v_{\text{probe}} - v_{\text{oracle}}| \le \tau$, where $\tau$ is a per-field tolerance declared in the game's perception specification. For angular fields (e.g., arc angles, ball rotation), circular distance $\min(d,\; 2\pi - d)$ is used to handle wrap-around.
  \item \textbf{Boolean}: Both sides are coerced to \texttt{bool} and compared by exact equality.
  \item \textbf{Enum / string}: Exact equality is required.
  \item \textbf{List by index} (e.g., \texttt{pipes}, \texttt{platforms}): Elements are matched positionally. Each element's sub-fields are compared via the rules above. If the probe returns fewer elements than the oracle, missing elements count as misses for all their expected sub-fields.
  \item \textbf{Nested list} (e.g., board grids in 2048, Minesweeper, Tetris): Cell-wise comparison---each cell $[r][c]$ is compared individually, and the score reflects the fraction of correctly identified cells.
  \item \textbf{List by color} (e.g., arcs in Color Switch): Elements are matched by a color key, then non-color sub-fields are compared via tolerance.
\end{itemize}

If the VLM's \texttt{perceive()} throws an exception on a frame, all expected attributes for that frame are counted as misses.

\medskip
\noindent\textbf{Tolerance calibration.}
Tolerances are set per game and per field by the benchmark authors based on the visual resolution and gameplay sensitivity of each attribute. Representative tolerances are shown in Table~\ref{tab:tolerances}. Spatial tolerances (in pixels) reflect the minimum detection error that would causally affect gameplay; tighter tolerances are used for grid-aligned games (e.g., Snake, Minesweeper) where cell-level accuracy is required, while looser tolerances apply to fast-scrolling games (e.g., Geometry Dash) where approximate obstacle positions suffice for control.

\begin{table}[t]
\centering
\small
\setlength{\tabcolsep}{3pt}
\begin{tabular}{@{}lllr@{}}
\toprule
\textbf{Game} & \textbf{Attribute} & \textbf{Tol.} & \textbf{Type} \\
\midrule
Flappy Bird
  & Bird.cx / .cy       & 8 px  & Position \\
  & Bird.velocity        & 0.5   & Physics \\
  & Pipe.gap\_top        & 25 px & Loose \\
  & score                & 1     & Exact \\
\midrule
Snake
  & Segment.x / .y      & 1 cell & Grid \\
  & score                & 1      & Exact \\
  & direction            & ---    & Enum \\
\midrule
Tetris
  & board cells          & ---    & Bool grid \\
  & cell\_size           & 2 px   & Const \\
  & grid\_w / h          & 0      & Exact \\
\midrule
2048
  & board cells          & ---    & Int grid \\
  & cell\_size           & 2 px   & Const \\
  & canvas\_width        & 3 px   & Const \\
\midrule
Geometry Dash
  & player\_y            & 10 px  & Scroll \\
  & obstacle.dist\_x     & 15 px  & Coarse \\
  & is\_grounded         & ---    & Bool \\
\midrule
Color Switch
  & Arc.start\_angle     & 0.25 rad & Circ. \\
  & Ball.color           & ---      & Enum \\
  & Circle.r             & 12 px    & Radius \\
\midrule
Knife Hit
  & Knife.angle          & 0.15 rad & Angular \\
  & knife\_speed         & 10\%     & Rel. \\
\midrule
Doodle Jump
  & Platform.type        & ---      & Enum \\
  & Player.velocity      & 2.0      & Physics \\
\bottomrule
\end{tabular}
\caption{Representative perception tolerances. ``---'' = exact match; ``Rel.'' = relative tolerance ($|v_p - v_o|/|v_o| \le \tau$). Full specs in \texttt{perception\_spec.py}.}
\label{tab:tolerances}
\end{table}

\subsubsection{API Score}

Stage 2 score measures functional correctness of the VLM-authored \texttt{GameAPI} class, which wraps perceptual state into strategy-facing methods (e.g., \texttt{get\_empty\_cells()}, \texttt{get\_closest\_obstacle()}). For each method on each frame, the grader invokes both the VLM's API (fed with the VLM's own Stage~1 probe output) and the oracle API (fed with ground-truth state), then compares their return values.

\medskip
\noindent\textbf{Per-method comparison strategies.}
Different return types require different comparison semantics. The grader dispatches to one of the following strategies per method:

\begin{itemize}
  \item \textbf{Exact equality} (\texttt{exact}): Python \texttt{==}---used for scalar returns, booleans, and simple structs.
  \item \textbf{Set equality} (\texttt{set\_equal}): Order-independent comparison---used for unordered collections (e.g., palette tuples).
  \item \textbf{Ordered equality} (\texttt{ordered\_equal}): Ordered list comparison.
  \item \textbf{Element-wise} (\texttt{by\_index}): Positional list comparison, used for method returns that are ordered sequences.
  \item \textbf{Tolerance} (\texttt{tolerance}): Absolute tolerance on numeric returns, analogous to Per.'s numeric comparison.
  \item \textbf{Dict tolerance} (\texttt{dict\_tol} / \texttt{dict\_tol\_rel}): Per-key comparison for dict returns, supporting both absolute and relative tolerances on individual keys.
\end{itemize}

The per-game API. is:
\[
\mathrm{API}_g = \frac{\text{method calls returning correct output}}{\text{total method calls}}.
\]

\medskip
\noindent\textbf{Diagnostic metric.}
In addition to the primary Stage 2 score (VLM API on VLM probe state), we report a secondary \emph{API-on-GT} metric: the VLM API evaluated on oracle ground-truth state. This isolates API implementation quality from upstream perception errors, allowing diagnosis of whether Stage~2 failure stems from incorrect API logic or from corrupted input state.

\subsubsection{Composite Metrics}

\medskip
\noindent\textbf{Obs.}
Obs. is the arithmetic mean of Per. and API. scores for a given game:
\[
\mathrm{Obs}_g = \frac{\mathrm{Per}_g + \mathrm{API}_g}{2}.
\]

\medskip
\noindent\textbf{Strategy Score (normalized).}
Raw Stage~3 strategy scores (game points) are not directly comparable across games. We apply per-game min-max normalization across all evaluated models:
\[
\mathrm{Stra}_g = \frac{s_g - \min_m s_{g,m}}{\max_m s_{g,m} - \min_m s_{g,m}},
\]
where $s_{g,m}$ is the raw mean score of model $m$ on game $g$ over 10 random seeds. When all models achieve the same score ($\max = \min$), the normalized score is $0$.

\medskip
\noindent\textbf{Joint Score.}
The end-to-end Joint score captures the multiplicative dependency between observation and strategy:
\[
\mathrm{Joint} = \frac{1}{G}\sum_{g=1}^{G} \mathrm{Obs}_g \times \mathrm{Stra}_g.
\]
A model that perceives perfectly but cannot act, or vice versa, receives a Joint score near zero---enforcing that both components must succeed for a model to score well.

\medskip
\noindent\textbf{Why Oracle-conditioned Strategy is the primary metric.}
A naively end-to-end metric---replacing Oracle perception with the VLM's own Stage~1/2 artifacts in Stage~3---collapses to near zero for most models (Table~\ref{tab:pipeline_ablation}, ``w/o Oracle'': 0.013 for GPT-5.6 Luna, 0.052 for Qwen 3.6 Plus). This is not informative: an error in any single perceived attribute (a mislocalized player, a misclassified obstacle) typically breaks the \texttt{play()} controller outright, so the resulting score conflates \emph{any} perception imperfection with total strategy failure. With most models clustered near zero, the metric loses discriminative power both across models and across games. We therefore report Oracle-conditioned Strategy as the primary metric (it isolates strategic reasoning once perception is held fixed) and retain the VLM-probe mode as a diagnostic in Table~\ref{tab:pipeline_ablation} for measuring how perception errors propagate into strategy collapse.

\subsection{Stage~0 Free-Exploration Mechanism}
\label{sec:stage0}

Stage~0 (Explore) provides the VLM with qualitative familiarity with the game before downstream stages attempt structured perception and control. The paper's phrase ``VLM freely plays the game for a short rollout'' is implemented as a \emph{code-generation-then-execution} pipeline rather than an interactive agent loop, as detailed below.

\subsubsection{Script Generation}

The VLM receives a text-only prompt (no screenshots) consisting of the game's natural-language rules and the sandbox environment API. It is asked to produce a single Python function \texttt{play(env)} that exercises the game. The available API is:

\begin{itemize}
  \item \texttt{env.reset()} --- reset the game to its initial state.
  \item \texttt{env.step(action\_id)} --- submit a discrete keyboard event (mapped to the game's configured key bindings via Playwright).
  \item \texttt{env.click(x, y)} --- issue a mouse click at pixel coordinates (for mouse-controlled games).
  \item \texttt{env.screenshot(label)} --- capture a labeled RGB frame for later analysis.
  \item \texttt{env.get\_score()} --- read the current visible game score.
  \item \texttt{env.mark\_game\_started()} --- signal that intentional play has begun (disables start-screen detection).
\end{itemize}

The VLM generates this script in a single forward pass; there is no iterative action-selection loop. If the generated code contains a syntax error, the grader retries up to two times with the error message appended to the prompt.

\subsubsection{Script Execution and Frame Capture}

The generated \texttt{play(env)} function is executed in a headless Chromium browser via Playwright. Each call to \texttt{env.step()} triggers the corresponding keyboard event on the game's DOM or Canvas element and automatically captures the resulting frame as an RGB numpy array. Labeled screenshots requested via \texttt{env.screenshot()} are saved separately. The script runs under a 90-second wall-clock timeout; execution also terminates on game-over or when the script returns.

After execution, a subset of frames is persisted to disk: the first, last, and evenly spaced intermediate frames (key frames).

\subsubsection{Analyst Distillation}

A second VLM call---the \emph{analyst}---receives the game rules, the exploration script, stdout output, and up to eight labeled screenshots from the rollout. The analyst is instructed to produce a structured markdown summary covering:

\begin{itemize}
  \item \textbf{Game phases}: start overlay, gameplay goal, game-over triggers.
  \item \textbf{Action effects}: one-line descriptions of visible changes per action.
  \item \textbf{Visual layout}: key objects and approximate spatial relationships (qualitative, no pixel estimates).
  \item \textbf{Dynamics}: objects that move, scroll, or change over time.
  \item \textbf{Non-obvious facts}: game mechanics that a downstream planner might misinterpret.
\end{itemize}

The analyst is explicitly instructed \emph{not} to estimate pixel coordinates, radii, or color ranges---those measurements are the responsibility of Stage~1.

\subsection{VLM-in-Loop Baseline: GameWorld-Style Continuous Inference}
\label{sec:ouyangGameWorldStandardizedVerifiable2026_loop}

To contextualize PlayScriptBench's batched pipeline against the dominant agent-loop paradigm, we implement a \emph{GameWorld-style continuous observation-action loop} as a baseline. This baseline mirrors the architecture used by GameWorld~\cite{ouyangGameWorldStandardizedVerifiable2026} and similar VLM agent benchmarks: at every timestep, the VLM receives a screenshot of the current game state and outputs a single action.

\subsubsection{Loop Architecture}

The loop follows a strict observe$\to$infer$\to$act cycle with no code generation or caching:

\begin{enumerate}
  \item \textbf{Pause and screenshot}: The game is paused, and the current frame is captured as an RGB image.
  \item \textbf{VLM inference}: The frame is base64-encoded and sent to the VLM along with a prompt containing the game rules, available actions, and a history of the last 5 steps. The model is instructed to reply with a single action name (e.g., \texttt{"left"}).
  \item \textbf{Resume and execute}: The game is resumed, the parsed action is mapped to a keyboard event and applied to the game via Playwright.
\end{enumerate}

Unlike PlayScriptBench's pipeline, no code is generated---the VLM is queried at every step, and its text output is directly interpreted as an action. There is no perceive function, no API abstraction, and no strategy search.

\subsubsection{Per-Step Prompt}

Each VLM call receives the following prompt structure:

\begin{itemize}
  \item \textbf{Role}: \texttt{You are playing a video game. Analyze the screenshot and choose the best action.}
  \item \textbf{Game rules}: natural-language description of the game (identical to the rules used in Stage~0).
  \item \textbf{Action space}: enumerated list of valid actions with their IDs (e.g., \texttt{0 (action up)}, \texttt{1 (action right)}).
  \item \textbf{History}: the last 5 steps with their chosen actions and resulting scores, providing short-term context.
  \item \textbf{Instruction}: \texttt{Reply with ONLY the action name (e.g. "right" or "jump"). No explanation.}
\end{itemize}

The system prompt is \texttt{You are a game player. Output only the action name.} No chain-of-thought, no structured output format, and no reasoning instructions are provided.

\subsubsection{Action Parsing}

The VLM's free-form text response is parsed by matching against known action names (case-insensitive substring match). If no match is found, a fallback dictionary maps common synonyms (e.g., \texttt{"jump"}$\to$\texttt{up}, \texttt{"move\_left"}$\to$\texttt{left}). On complete parse failure, the first action in the action space is selected as default.

\subsubsection{Metrics Collection}

Every VLM call is instrumented via the shared \texttt{MetricsCollector}, recording prompt tokens, completion tokens, total tokens, reasoning tokens, and wall-clock latency. The per-step log captures the action taken, score, reward, and whether the game terminated, enabling post-hoc analysis of token consumption and agent behavior.

\subsection{Pipeline Configuration: Prompt Templates and Feedback Formats}
\label{sec:pipeline_config}

This appendix details the prompt structure and inter-stage feedback mechanisms for each stage of the four-stage pipeline.

\subsubsection{Stage~0: Free Exploration}

Stage~0 runs for a single round. The round proceeds in three phases: the VLM generates a \texttt{play(env)} exploration script, the script is executed under a 90-second wall-clock timeout, and a VLM analyst distills captured screenshots and stdout into structured insights. Syntax errors in the generated script trigger up to 2 retries with the error message appended.

The script-generation prompt carries the system message \texttt{You are a game exploration agent. Output a Python play(env) function.} and includes the game rules, the environment API documentation (\texttt{env.reset()}, \texttt{env.step(action)}, \texttt{env.screenshot(label)}, \texttt{env.get\_score()}, \texttt{env.click(x,y)}, \texttt{env.mark\_game\_started()}), and instructions to freely operate the game while taking screenshots before and after each action. The prompt explicitly forbids using \texttt{env.annotate()}, printing pixel coordinates or color values, and attempting to score well.

The analyst prompt receives the game rules, the exploration script (truncated to 2000 characters), stdout (truncated to 1500 characters), and up to 8 labeled screenshots. Under the system message \texttt{You are a careful game analyst. Output markdown bullet points.}, the analyst must produce insights under five headings: Game Phases, Action Effects, Visual Layout, Dynamics, and Non-obvious Facts, without estimating pixel coordinates or quantitative measurements. The final artifact is a consolidated markdown file passed to Stage~1 as qualitative context.

\subsubsection{Stage~1: Perception}

Stage~1 runs for a single round with up to 2 syntax retries. The exam prompt uses the system message \texttt{You are a computer vision expert. Output a single Python function.} and presents the game rules, any exploration insights from Stage~0, the complete \texttt{perception\_spec.py} source verbatim (defining the \texttt{GameState} dataclass and \texttt{perceive()} stub), and, when applicable, constants calibration samples as a table of experimental measurements. The task is to complete \texttt{perceive(frame)} (or \texttt{perceive(frame, prev\_frame)} for two-frame games) and optionally fill in \texttt{CONSTANTS = Constants(...)} with concrete values. The prompt emphasizes that the transformation must be pure (no env calls, no I/O) and that the dataclasses are pre-defined and must not be re-declared. It is accompanied by 3 diverse sample frames biased toward interesting non-empty states.

The prompt includes a self-verification section in which the probe is executed on up to 5 evenly-spaced frames and a per-field textual summary is produced for each, allowing the VLM to visually compare its detection values against the images.

\subsubsection{Stage~2: API Construction}

Stage~2 runs for a single round with early stopping at 95\% API accuracy and up to 2 syntax retries. The prompt uses the system message \texttt{You are a Python expert.} and provides the game rules, the complete \texttt{perception\_spec.py} source (so the VLM understands the state shape received by the constructor), the complete \texttt{api\_spec.py} source (containing the \texttt{GameAPI} class template and \texttt{METHOD\_SPECS} with method signatures, return types, and comparison strategies), and sample frame count for visual context. The VLM must implement only the \texttt{GameAPI} class body, where the constructor receives a \texttt{SimpleNamespace} and the \texttt{get\_*} methods must be pure (no env calls, no I/O, no derived quantities or decision logic). When Stage~1 is skipped, a variant prompt is used in which the constructor receives a raw \texttt{RGB ndarray} instead of perceived state, and the VLM must implement perception logic inside the class methods.

\subsubsection{Stage~3: Strategy Composition}

Stage~3 runs for a single round with up to 3 syntax retries and a 60-second wall-clock timeout per \texttt{play()} execution. Verification runs the generated script with 3 screening seeds (\texttt{[42, 123, 456]}) and additional robustness seeds.

The prompt uses the system message \texttt{Write a Python game automation script using the provided GameAPI.} and includes the game rules, available actions, environment API documentation, calibration constants as \texttt{UPPERCASE} module-level variables, GameAPI method signatures (docstrings only, not full implementation, to prevent copying stubs), For games requiring real-time measurement of physical constants (e.g., \texttt{GRAVITY}, \texttt{TAP\_IMPULSE}, \texttt{EXTEND\_RATE}), a physics probe section instructs the VLM to measure these via environment experiments at the start of \texttt{play()}. The prompt constrains that \texttt{perceive()} and \texttt{GameAPI} are in scope and must not be redefined, a new \texttt{GameAPI} instance must be created on every frame, the function must exit on \texttt{terminated=True}, and \texttt{time.sleep()} is forbidden.

\subsubsection{Inter-Stage Feedback Summary}

Table~\ref{tab:feedback_summary} summarizes the information passed between stages.

\begin{table*}[t]
\centering
\small
\setlength{\tabcolsep}{4pt}
\begin{tabular}{@{}p{2.0cm}p{3.4cm}p{2.0cm}p{5.2cm}@{}}
\toprule
\textbf{Source} & \textbf{Content Passed} & \textbf{Exposed to VLM?} & \textbf{Format} \\
\midrule
Stage~0 $\to$ 1 & Exploration insights & Yes & Markdown bullet points (game phases, action effects, visual layout, dynamics, non-obvious facts) \\
\midrule
Stage~1 internal & Self-verify output & Yes (no grader scores) & Textual detection summary + raw frames; syntax/runtime errors \\
\midrule
Stage~1 $\to$ 2 & Best probe.py + constants & Yes (executed into scope) & Python source code + JSON constants \\
\midrule
Stage~2 internal & Previous code + crash trace & Yes (no scoring) & Full code + first error (600 chars) \\
\midrule
Stage~2 $\to$ 3 & Best api.py & Yes (executed into scope) & Python source code \\
\midrule
Stage~3 internal & Previous results & Yes & Score, survival time, status, stdout (3000 chars), error log (2000 chars), previous script \\
\bottomrule
\end{tabular}
\caption{Information flow between stages. Grader scores are deliberately withheld from VLM feedback in Stages~1 and~2 to force self-correction via visual inspection and error analysis.}
\label{tab:feedback_summary}
\end{table*}

\subsection{Perception--Strategy Correlation Analysis}
\label{sec:spearman}

To quantify the relationship between visual grounding and strategic reasoning, we compute the Spearman rank correlation between overall Obs. and Strategy scores across all 13 models. The correlation is weak and not statistically significant ($\rho = 0.340$, $p = 0.255$), confirming that Observation and Strategy capture largely independent capabilities: a model that extracts accurate visual state does not necessarily produce effective gameplay, and vice versa. Table~\ref{tab:rank_inversions} reports the per-model ranks. The most extreme inversions are GPT-5.6 Luna (Obs.\ rank 1, Strategy rank 11; $\Delta = +10$) and Gemini 3.5 Flash (Obs.\ rank 13, Strategy rank 6; $\Delta = -7$), demonstrating that strong perception without strategic reasoning, and strong reasoning from weak perception, are both common patterns. These results confirm that the Observation--Strategy decomposition is not merely a formal separation but captures genuinely distinct capability dimensions.

\begin{table}[tb]
\centering
\small
\setlength{\tabcolsep}{2pt}
\begin{tabular}{@{}lccccc@{}}
\toprule
\textbf{Model} & \textbf{Obs.} & \textbf{Stra.} & \textbf{Rk$_{\mathrm{Obs}}$} & \textbf{Rk$_{\mathrm{Stra}}$} & \textbf{$\Delta$} \\
\midrule
GPT-5.6 Luna         & 0.53 & 0.255 & 1  & 11 & $+10$ \\
Claude Sonnet 4.6    & 0.53 & 0.414 & 1  & 4  & $+3$  \\
MiniMax-M3           & 0.52 & 0.345 & 3  & 8  & $+5$  \\
Doubao Seed 2.1 Turbo& 0.51 & 0.427 & 4  & 3  & $-1$  \\
GPT-5.4 Mini         & 0.51 & 0.343 & 4  & 9  & $+5$  \\
Kimi K2.5            & 0.51 & 0.459 & 4  & 2  & $-2$  \\
Qwen 3.6 Plus        & 0.47 & 0.600 & 7  & 1  & $-6$  \\
Claude Sonnet 5      & 0.46 & 0.292 & 8  & 10 & $+2$  \\
ERNIE 5.0            & 0.45 & 0.392 & 9  & 5  & $-4$  \\
Qwen 3.5 Flash       & 0.44 & 0.277 & 10 & 12 & $+2$  \\
Gemini 3.1 Flash Lite& 0.39 & 0.233 & 11 & 13 & $+2$  \\
GLM-4.6V             & 0.36 & 0.144 & 12 & 7  & $-5$  \\
Gemini 3.5 Flash     & 0.34 & 0.360 & 13 & 6  & $-7$  \\
\midrule
\multicolumn{3}{l}{Spearman $\rho$} & \multicolumn{2}{c}{0.340} & ($p{=}0.255$) \\
\bottomrule
\end{tabular}
\caption{Per-model Observation and Strategy ranks (rank 1 = highest score). $\Delta = \mathrm{Rk}_{\mathrm{Stra}} - \mathrm{Rk}_{\mathrm{Obs}}$; positive values indicate overperformance in Strategy relative to Observation.}
\label{tab:rank_inversions}
\end{table}

\subsection{Failure Mode Analysis}
\label{sec:diagnostic}

We categorize model failures into three recurring patterns across the pipeline, each tied to a distinct cognitive limitation.

\textbf{(P1)~Hallucinated visual grounding (Stage~1).} Rather than leveraging the structured \texttt{perceive()} interface, models frequently author hand-coded computer-vision pipelines that guess visual parameters from scratch. A representative example is HSV color thresholding: models hallucinate hue/saturation/value ranges (e.g., \texttt{lower\_yellow = [20, 100, 100]}) that may not match the actual game sprites---producing zero detection accuracy despite the game having high-contrast visual elements. More subtly, models occasionally apply color-space conversions incorrectly (e.g., BGR2HSV on RGB input), producing systematically shifted detections. This failure pattern is widespread: across the evaluated models, few consistently avoid inventing detection heuristics when the prompt provides the opportunity to do so.

\textbf{(P2)~Perception--API cascade (Stage~2).} Stage~2 failures are dominated by upstream error propagation rather than API code defects. When Stage~1 perception returns partial or invalid state, the \texttt{GameAPI} constructor receives a \texttt{None}-typed state object, causing \texttt{AttributeError} on every method invocation. On Doodle Jump, GLM-4.6V achieves 0.85 perception accuracy but scores 0.0 on API construction because the \texttt{perceive()} function fails on boundary frames (game-over screens), injecting \texttt{None} into the state stream. Even when perception is structurally correct, CV-derived coordinates contain pixel-level noise (off by 1--2~px per frame), which accumulates across stacked method calls and degrades API accuracy below the threshold needed for downstream strategy. The cascade effect means that a model's Stage~2 score is bounded above by its Stage~1 accuracy---a ceiling that cannot be broken through without improving perception first.

\textbf{(P3)~Syntactically valid, strategically vacuous (Stage~3).} The most common Stage~3 failure is code that parses and executes without error but produces no meaningful gameplay. Among all Stage~3 rounds, a larger class of failures produces syntactically valid controllers with flawed decision logic: Snake controllers use greedy single-step pathfinding without accounting for body growth, and Tetris controllers lack piece-rotation awareness. These scripts run for hundreds of frames, consume full time budgets, yet score zero because the reasoning---not the execution---is defective. This pattern is particularly diagnostic: it separates models that can write runnable code from those that can reason about sequential decision-making under uncertainty.

\subsection{Implementation Details}
\label{sec:implementation}

\subsubsection{Game Origins}

The 16 benchmark games are sourced from three categories to balance variety with reproducibility:

\begin{itemize}
  \item \textbf{Self-developed} (5 games): Color Switch, Knife Hit, Stick Hero, Stack2D, and Match-3. These are original HTML5 Canvas implementations authored by the benchmark authors, with deterministic physics constants and no external dependencies.
  \item \textbf{Adapted from GameWorld} (3 games): Chrome Dino, Minesweeper, and Tetris. These were adapted from the GameWorld~\cite{ouyangGameWorldStandardizedVerifiable2026} benchmark suite with modified environment wrappers for PlayScriptBench's sandbox interface. Tetris uses the open-source TwitchTetris HTML5 implementation.
  \item \textbf{Open-source / community games} (8 games): Space Invaders (MIT, Susam Pal), Doodle Jump, Flappy Bird, Geometry Dash, Helix Jump, Pac-Man, Snake, and 2048. These are sourced from the Offline-HTML-Games-Pack~\cite{offlinehtmlgames} community collection or standalone repositories, wrapped with the standardized environment interface.
\end{itemize}

All games run as client-side HTML5 Canvas applications in a headless Chromium browser via Playwright ($\ge$ 1.40.0). No server-side logic or network access is required.

\subsubsection{Observation Resolution}

Each game specifies a fixed viewport size that determines the resolution of screenshots captured by the sandbox. The Playwright browser context is created with \texttt{device\_scale\_factor=1}, ensuring no device-pixel-ratio scaling. Screenshots are captured as JPEG (quality 80) from the game's \texttt{<canvas>} element (or the full page when no canvas is present) and resized to the canonical resolution via bilinear interpolation. Table~\ref{tab:viewport} lists the per-game viewport dimensions.

\begin{table}[tb]
\centering
\small
\setlength{\tabcolsep}{3pt}
\begin{tabular}{@{}ll@{}ll@{}}
\toprule
\textbf{Game} & \textbf{Resolution} & \textbf{Game} & \textbf{Resolution} \\
\midrule
Flappy Bird     & $640{\times}480$  & Geometry Dash  & $960{\times}540$  \\
Chrome Dino     & $600{\times}150$  & 2048           & $500{\times}650$  \\
Snake           & $420{\times}460$  & Minesweeper    & $500{\times}600$  \\
Pac-Man         & $560{\times}600$  & Doodle Jump    & $422{\times}552$  \\
Tetris          & $600{\times}500$  & Helix Jump     & $400{\times}600$  \\
Color Switch    & $500{\times}700$  & Knife Hit      & $500{\times}700$  \\
Stick Hero      & $500{\times}700$  & Stack2D        & $500{\times}700$  \\
Space Invaders  & $640{\times}480$  & Match-3        & $480{\times}720$  \\
\bottomrule
\end{tabular}
\caption{Per-game observation viewport (pixels). All screenshots are captured at these resolutions via Playwright with $\textit{device\_scale\_factor}=1$.}
\label{tab:viewport}
\end{table}

\subsubsection{Action Space}

Each game defines a discrete action space mapped to keyboard events or mouse inputs via Playwright. Actions are integer-indexed starting from 0, where action 0 is universally ``do nothing'' (no input sent). The action-to-key mapping is declared per-game in the environment configuration (\texttt{config.py} or \texttt{game.yaml}) and injected into the Stage~3 prompt so the VLM knows the exact action vocabulary. Table~\ref{tab:action_spaces} summarizes the action spaces by category.

\begin{table}[tb]
\centering
\small
\setlength{\tabcolsep}{0.5pt}
\begin{tabular}{@{}p{1.6cm}p{1.0cm}p{5.7cm}@{}}
\toprule
\textbf{Category} & \textbf{$|\mathcal{A}|$} & \textbf{Games and actions} \\
\midrule
4-directional  & 4 & 2048, Pac-Man, Snake: up / right / down / left \\
\midrule
Binary         & 2 & Flappy Bird: do\_nothing / flap (Space); Color Switch: do\_nothing / tap; Knife Hit: do\_nothing / throw; Stack2D: do\_nothing / place; Geometry Dash: do\_nothing / jump \\
\midrule
3-action       & 3 & Chrome Dino: do\_nothing / jump / duck; Doodle Jump: do\_nothing / left / right; Helix Jump: do\_nothing / rotate\_left / rotate\_right \\
\midrule
Multi-action   & 4--7 & Space Invaders: do\_nothing / left / right / fire; Tetris: left / right / rotate\_cw / rotate\_ccw / soft\_drop / hard\_drop / hold \\
\midrule
Continuous     & 20 & Stick Hero: do\_nothing / 19 press durations (100--1000\,ms, 50\,ms increments) \\
\midrule
Grid-click     & 65 & Match-3: do\_nothing / click on $8{\times}8$ grid cell; Minesweeper: click / flag / chord (mouse-based) \\
\bottomrule
\end{tabular}
\caption{Action space summary. $|\mathcal{A}|$ denotes the number of discrete actions. Mouse-based games (Match-3, Minesweeper, Stick Hero) use pixel-coordinate or duration mappings in addition to key bindings.}
\label{tab:action_spaces}
\end{table}

\subsubsection{Sandbox Timing and Frame Control}

The sandbox does not impose a fixed global frame rate. Instead, game progression is controlled by a per-game \texttt{step\_delay\_ms} parameter---the wall-clock wait time injected between consecutive action steps via \texttt{page.wait\_for\_timeout()}. This design aligns the sandbox step interval with each game's native frame time, ensuring that real-time games (e.g., Geometry Dash at 33\,ms $\approx$ 30\,FPS) advance at approximately their intended speed, while turn-based games (e.g., 2048, Minesweeper) use longer delays (100\,ms) to allow animation completion.

Between action steps, the sandbox \emph{pauses} the game loop by monkey-patching \texttt{requestAnimationFrame}, \texttt{setTimeout}, \texttt{setInterval}, \texttt{Date.now}, and \texttt{performance.now} in the browser context. When paused, all pending animation-frame callbacks are queued and replayed upon resume. This freeze-during-inference mechanism ensures that every agent faces identical game dynamics regardless of response latency---the game clock advances only during active step intervals.

Representative step delays: 30\,ms (Color Switch), 33\,ms (Flappy Bird, Geometry Dash, Space Invaders), 50\,ms (Snake, Chrome Dino, Tetris), 80\,ms (Knife Hit), and 100\,ms (2048, Minesweeper, Doodle Jump). Match-3 uses a special mode with 0\,ms step delay and a 1500\,ms post-step delay to allow match animations to complete.

\end{document}